\documentclass[10pt,twocolumn,letterpaper]{article}

\usepackage{cvpr}              
\usepackage{algorithm} 
\usepackage{algpseudocode} 
\usepackage{amsmath}
\usepackage{soul}
\usepackage{booktabs}
\usepackage{overpic}
\usepackage{amssymb}
\usepackage{stfloats}
\usepackage[accsupp]{axessibility}
\usepackage{pifont}
\usepackage{multirow}

\usepackage[dvipsnames]{xcolor}

\definecolor{cvprblue}{rgb}{0.21,0.49,0.74}
\usepackage[pagebackref,breaklinks,colorlinks,citecolor=cvprblue]{hyperref}

\def\paperID{9907} 
\def\confName{CVPR}
\def\confYear{2024}

\title{MaskClustering:  View Consensus based Mask Graph Clustering\\ for Open-Vocabulary 3D Instance Segmentation}

\author{
Mi Yan\textsuperscript{1,2} \quad 
Jiazhao Zhang\textsuperscript{1,2} \quad 
Yan Zhu\textsuperscript{1} \quad 
He Wang\textsuperscript{1,2,3,$^\dagger$} \\
\textsuperscript{1}CFCS, School of CS, Peking University \quad 
\textsuperscript{2}Beijing Academy of Artificial Intelligence 
\textsuperscript{3}Galbot \quad \\
}

\begin{document}

\twocolumn[{%
\renewcommand\twocolumn[1][]{#1}%
\maketitle
\vspace{-1cm}
\begin{center}
    \centering
    \captionsetup{type=figure}
    \includegraphics[width=1\linewidth]{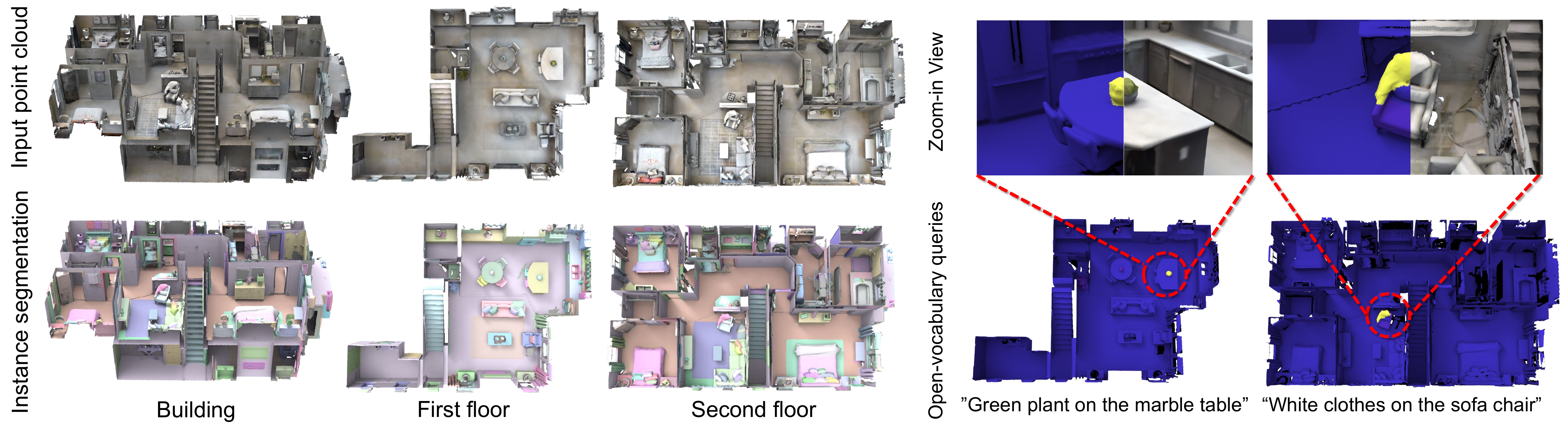}
    \captionof{figure}{Our method tackles the challenges of open-vocabulary instance segmentation. It achieves detailed segmentation across objects of varying scales and can query these objects using open-vocabulary text.}
    \label{fig:teaser}
\end{center}
}]

\maketitle

\begin{abstract}

\let\thefootnote\relax\footnote{$^\dagger$: He Wang is the corresponding author.}

Open-vocabulary 3D instance segmentation is cutting-edge for its ability to segment 3D instances without predefined categories. However, progress in 3D lags behind its 2D counterpart due to limited annotated 3D data. To address this, recent works first generate 2D open-vocabulary masks through 2D models and then merge them into 3D instances based on metrics calculated between two neighboring frames. In contrast to these local metrics, we propose a novel metric, view consensus rate, to enhance the utilization of multi-view observations. The key insight is that two 2D masks should be deemed part of the same 3D instance if a significant number of other 2D masks from different views contain both these two masks. Using this metric as edge weight, we construct a global mask graph where each mask is a node. Through iterative clustering of masks showing high view consensus, we generate a series of clusters, each representing a distinct 3D instance. Notably, our model is training-free. Through extensive experiments on publicly available datasets, including ScanNet++, ScanNet200 and MatterPort3D, we demonstrate that our method achieves state-of-the-art performance in open-vocabulary 3D instance segmentation. Our project page is at \href{https://pku-epic.github.io/MaskClustering/}{https://pku-epic.github.io/MaskClustering}.

\end{abstract}   
\section{Introduction}
\label{sec:intro}

Open-vocabulary 3D instance segmentation tackles the problem of predicting 3D object instance masks and their corresponding categories from reconstructed 3D scenes, without relying on a predefined list of categories. This is an essential task for 3D scene understanding~\cite{rozenberszki2022language, dai2017scannet, hou20193d}, robotics~\cite{Gu2023ConceptGraphsO3, zhang20233d, huang2023visual} and VR/AR applications~\cite{kerr2023lerf,  yamazaki2023open}. However, this task is more challenging than its established 2D counterpart, open-vocabulary 2D instance segmentation ~\cite{Wu2023BetrayedBC, Wang2023SAMCLIPMV, Vibashan2023MaskFreeOO, He2023SemanticPromotedDA, Huynh2021OpenVocabularyIS}, primarily due to the lack of large-scale open-world 3D data. Consequently, most current methods~\cite{Lu2023OVIR3DO3, takmaz2023openmask3d, Huang2023OpenIns3DSA} in this field divide this task into two stages: zero-shot 3D instance mask prediction, followed by open-vocabulary semantic queries. In this work, we primarily focuses on obtaining high-quality, zero-shot 3D instance masks.

Existing approaches for zero-shot 3D instance mask prediction primarily follow two paths.
3D-to-2D projection-based methods~\cite{huang2023openins3d, Huang2023OpenIns3DSA} leverage existing 3D instance segmentation algorithms to generate 3D masks. However, this approach is fundamentally constrained by the quality of 3D reconstructions and the relatively modest capabilities of current 3D instance segmentation tools. As a result, these methods often struggle to accurately segment small objects, leading to a significant loss of detail in complex scenes. In contrast, 2D-to-3D region grow-based methods~\cite{Lu2023OVIR3DO3, yang2023sam3d} leverage 2D segmentation models to process frames sequentially and update a list of 3D instances simultaneously. They merge new 2D masks with existing 3D instances based on geometric overlap and semantic similarity for each frame. However, we find that such online processing lacks global optimality across all frames, often resulting in incorrect merging.

To address these limitations, we propose a novel approach that improves global consistency via multi-view verification, inspired by bundle adjustment~\cite{triggs2000bundle}. Unlike prior methods that rely on local metrics calculated between adjacent frames to decide whether a mask pair should be merged, our method introduces a new global metric, the view consensus rate, which measures the proportion of frames supporting their merging. Here, a frame $t$ supports merging only if another 2D mask within frame $t$ contains this mask pair. In this way, the same-instance relationship of two view-consensus masks are indeed supported by multi-view observation. 

Utilizing the same-instance relationship, we build a global mask graph wherein each node is a mask, with edges added between high view consensus mask pairs. Following this, mask pairs exhibiting high view consensus are prioritized for merging into a mask cluster, and the view consensus between this mask cluster and other mask clusters will be updated. This iterative clustering and updating process yields a final list of clusters, each containing multiple masks and denoting a 3D instance. For each 3D instance, its point cloud and semantic feature are the aggregated partial point clouds and open-vocabulary features derived from individual 2D masks, respectively.

Our method, validated on ScanNet++ \cite{yeshwanth2023scannet++}, Matterport3D \cite{Matterport3D}, and ScanNet200 \cite{rozenberszki2022language} benchmarks, achieves state-of-the-art results in zero-shot mask prediction and open-vocabulary instance understanding, surpassing existing methods, especially in segmenting fine-grained objects.

Our contributions can be concluded as follows:
\begin{itemize}
\item A novel graph clustering based methodology to merge 2D masks for 3D open-vocabulary instance segmentation.
\item A novel view consensus metric for evaluating the relationship between 2D masks, effectively leveraging global information from input image sequences.
\item A SOTA open-vocabulary 3D instance segmentation method, which demonstrates superior performance on many publicly available datasets. 
\end{itemize}
\section{Related Works}
\label{sec:relatedworks}

\textbf{Closed-set 3D instance segmentation.}
Since the emergence of 3D scene datasets~\cite{He2023SemanticPromotedDA, dai2017scannet}, the computer vision community has witnessed a large literature of 3D segmentation methods~\cite{choy20194d, han2020occuseg, hu2021bidirectional, hu2021vmnet, robert2022learning, huang2019texturenet, li2022panoptic, schult2022mask3d, vu2022softgroup, vu2022softgroup++}. These methods tackle this problem either in online~\cite{liu2022ins, narita2019panopticfusion, zhang2020fusion, huang2021supervoxel,zheng2019active} or offline~\cite{schult2022mask3d, vu2022softgroup, vu2022softgroup++, robert2022learning} manner, representing the scene as points cloud, voxels, and more recently neural field~\cite{zhi2021place, vora2021nesf}. Though significant progress has been made, these methods are limited to a closed-set category list which is pre-defined in certain dataset, suffering poor performance in open-vocabulary settings as tail classes that have few or no training examples. In contrast, our method aims to tackle open-vocabulary 3D instance segmentation that segment objects of arbitrary category.

\noindent\textbf{Open-vocabulary 2D instance segmentation.}
The recent advances in large visual foundation models~\cite{kirillov2023segment, qi2023high, cheng2022masked,radford2021learning, Li2022LanguagedrivenSS, Ghiasi2021ScalingOI} have enabled a remarkable level of robustness of 2D understanding tasks. Typical tasks include zero-shot 2D segmentation~\cite{kirillov2023segment, cheng2022masked, qi2023high}, open-vocabulary 2D image understanding~\cite{radford2021learning, Li2022LanguagedrivenSS, Ghiasi2021ScalingOI}, and open-vocabulary 2D object detection~\cite{zhou2022detecting, kaul2023multi, kim2023region}. 
Recently, many works~\cite{Wu2023BetrayedBC, Wang2023SAMCLIPMV, Vibashan2023MaskFreeOO, He2023SemanticPromotedDA, Huynh2021OpenVocabularyIS} focus on the open-vocabulary 2D segmentation task, which requires predicting the open-vocabulary feature at the pixel level. These methods encode 2D images and align open-vocabulary pixel features with them. 
However, due to the lack of large-scale 3D annotated data, end-to-end open-vocabulary 3D instance segmentation is in slow progress. In this work, we tackle the open-vocabulary 3D instance segmentation by leveraging the prior from large 2D vision-language models.

\noindent\textbf{Open-vocabulary 3D instance segmentation.} There are two types of methods: (1) 3D-to-2D projection methods and (2) 2D-to-3D region grow-based methods.
(1) 3D-to-2D projection methods~\cite{takmaz2023openmask3d, huang2023openins3d, peng2023openscene} directly conduct 3D instance segmentation~\cite{schult2022mask3d, huang2023openins3d} on 3D indoor scene input. They project the 3D instance objects to 2D frames, and extract open-vocabulary features for final aggregation. However, these types of methods are limited to well-reconstructed scene and detailed objects are easily missed if the geometry details are poor. (2) 2D-to-3D region grow-based methods~\cite{Gu2023ConceptGraphsO3, Lu2023OVIR3DO3} propose to online fuse 2D observation to 3D instance segmentation. By back-projecting the 2D mask to 3D point cloud, these methods leverage clustering algorithm~\cite{ester1996density} or geometry overlapping to find corresponding 3D instances. The open-vocabulary feature is also aggregated during the back-projection. However, these types of methods consider the associations between historical constructed 3D instances with live frame, lacking a global understanding of all observed frames. 

Concurrently, SAI3D\cite{yin2023sai3d} and Open3DIS\cite{nguyen2023open3dis} propose merging 3D superpoints\cite{felzenszwalb2004efficient} guided by predictions from SAM\cite{kirillov2023segment}, showing robust performance in open-vocabulary 3D instance segmentation. However, we diverge from their approach by avoiding reliance on 3D superpoints, which face challenges in distinguishing geometrically-homogeneous objects like posters on walls or rows of similar medicine boxes.
\begin{figure*}[t]
\centering
\includegraphics[width=1\linewidth]{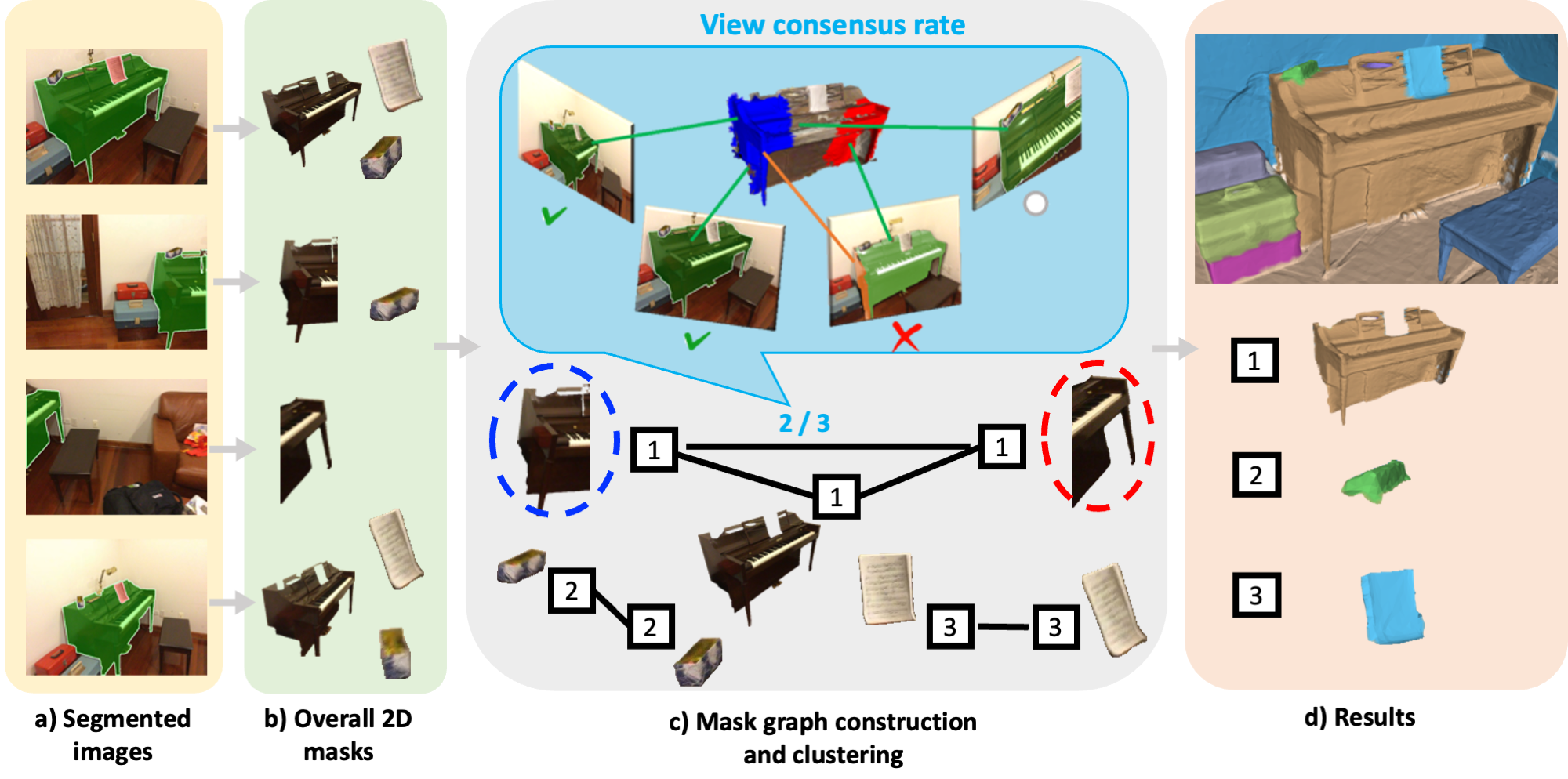}
\caption{
Overview pipeline of our method: a) We take segmented image sequences as input and b) extract all 2D masks from the input. c) To merge them, we build a global graph with each node as a mask. We use the view consensus rate, which is defined as the proportion of frames supporting the merging, to add edges between nodes. Each frame supports the merging only if there is a mask in this frame containing both nodes. d) Each mask cluster is merged into a 3D instance. For clarity, we only visualize three objects in the figure.
}
\label{fig:pipeline}
\vspace{-1mm}
\end{figure*}

\section{Method}

\subsection{Problem Formulation and Method Overview}
Given a set of posed color images $\{I_1^c, I_2^c, \ldots, I_T^c \}$, their corresponding depths $\{I_1^d, I_2^d, \ldots, I_T^d \}$, and the reconstructed point cloud $P$ of a scene, our algorithm outputs a list of 3D instances along with their open-vocabulary semantics fused from 2D mask proposals. 

We initially employ an off-the-shelf, class-agnostic mask predictor to process each color image $I_t^c$ and derive the 2D masks $\{m_{t,i}\mid i=1,2,\ldots,n_t\}$ where $n_t$ denotes the number of masks in frame $t$. We assume the mask predictor to generate entity-level panoptic segmentation masks, indicating that each mask approximates one object with nearly all pixels assigned to a single mask. This assumption aligns with capabilities of advanced segmentation tools like CropFormer\cite{qi2023high}.

The overview pipeline of our method is illustrated in Fig.~\ref{fig:pipeline}. To fuse these 2D masks from different frames into 3D instances, we propose to construct a mask graph $G = (V, E)$.  Each node in $V$ corresponds to a mask $m_{t,i}$, and an edge in $E$ indicates that two masks are part of the same instance and should be merged. To assess edge connectivity, we propose to leverage consensus cues from multi-view observations and therefore introduce view consensus rate as a criterion (Sec. \ref{graph_construction}).

Once the mask graph is established, we initiate an iterative process to cluster masks and update edges, with a priority on merging mask pairs displaying solid view consensus (Sec. \ref{iterative}). The result of this iterative process is a list of clusters, each denoting a 3D instance and containing multiple masks. Within such a cluster, we aggregate the corresponding partial point clouds from the individual masks to form the ultimate 3D instance. Building on these correspondences between 2D masks and 3D instances, we perform feature fusion for a more comprehensive representation, which aids in open-vocabulary semantic prediction (Sec. \ref{open-voc}).

\subsection{Mask Graph Construction}
In this subsesction, we introduce view consensus rate, which serves as the criterion to determine edge connectivity between two masks (Sec.\ref{view_consensus}). We then propose an efficient method for calculating this rate (Sec.\ref{efficient_consensus}) and leverage this rate to filter out under-segmented masks (Sec.\ref{mask_filter}).
\\
\noindent\textbf{Notations and Definitions}
Given the reconstructed point cloud $P$ and frame index $t$, for a mask $m_{t,i}$, we can obtain the mask point cloud  $P_{t,i}$ by projecting onto $P$ the backprojected point cloud of $m_{t,i}$ from $I_t^d$. Then we define the frame point cloud $P_{t}$ as the union of all $P_{t,i}$s for $i=1,2,...,n_t$, yielding $P_{t,i} \subset P_{t} \subset P$.
We define a point $p$ to be visible at frame $t$ if $p \in P_t$.
We then define a mask $m_{t',i}$ to be visible at frame $t$ if at least $\tau_{vis}=0.3$ of its total points from $P_{t',i}$ are visible and denote the visible part as $P_{t',i}^{t}$. We denote the set of frames where $m_{t',i}$ is visible as $F(m_{t',i})$.
Finally, we define the approximate containment relationship of one point clouds $P_i$ by another point cloud $P_j$ as $P_i\sqsubset P_j$, if at least $\tau_{contain}=0.8$ of the total points in $P_{i}$ lie within $P_{j}$.

\vspace{-3mm}

\label{graph_construction}
\subsubsection{View Consensus Rate}
\label{view_consensus}

The cornerstone of our method lies in determining if two masks belong to the same instance by utilizing 2D predictions across all frames. In this context, we propose to leverage view consensus cues, as detailed below.

To assess the relationship between two masks, specifically $m_{t',i}$ and $m_{t'',j}$, where $t'$ and $t''$ may be the same or different frames, we utilize the masks $\{m_{t,k}\}$ from relevant views. The goal is to check if there is substantial consensus among the views supporting that these two masks represent the same 3D instance.

To be more specific, we first find all the frames $O$ in which both of the two masks are visible, serving as the observers to the two masks, \textit{i.e.}, $O(m_{t',i}, m_{t'',j})= F(m_{t',i}) \cap F(m_{t'',j})$. And we denote the number of observers in $O$ as $n(m_{t',i}, m_{t'',j}) = \vert O(m_{t',i}, m_{t'',j})\vert$, where $| \cdot |$ represents the cardinality of the set.

\begin{figure}[t]
\centering
\begin{overpic}
[width=1\linewidth]
{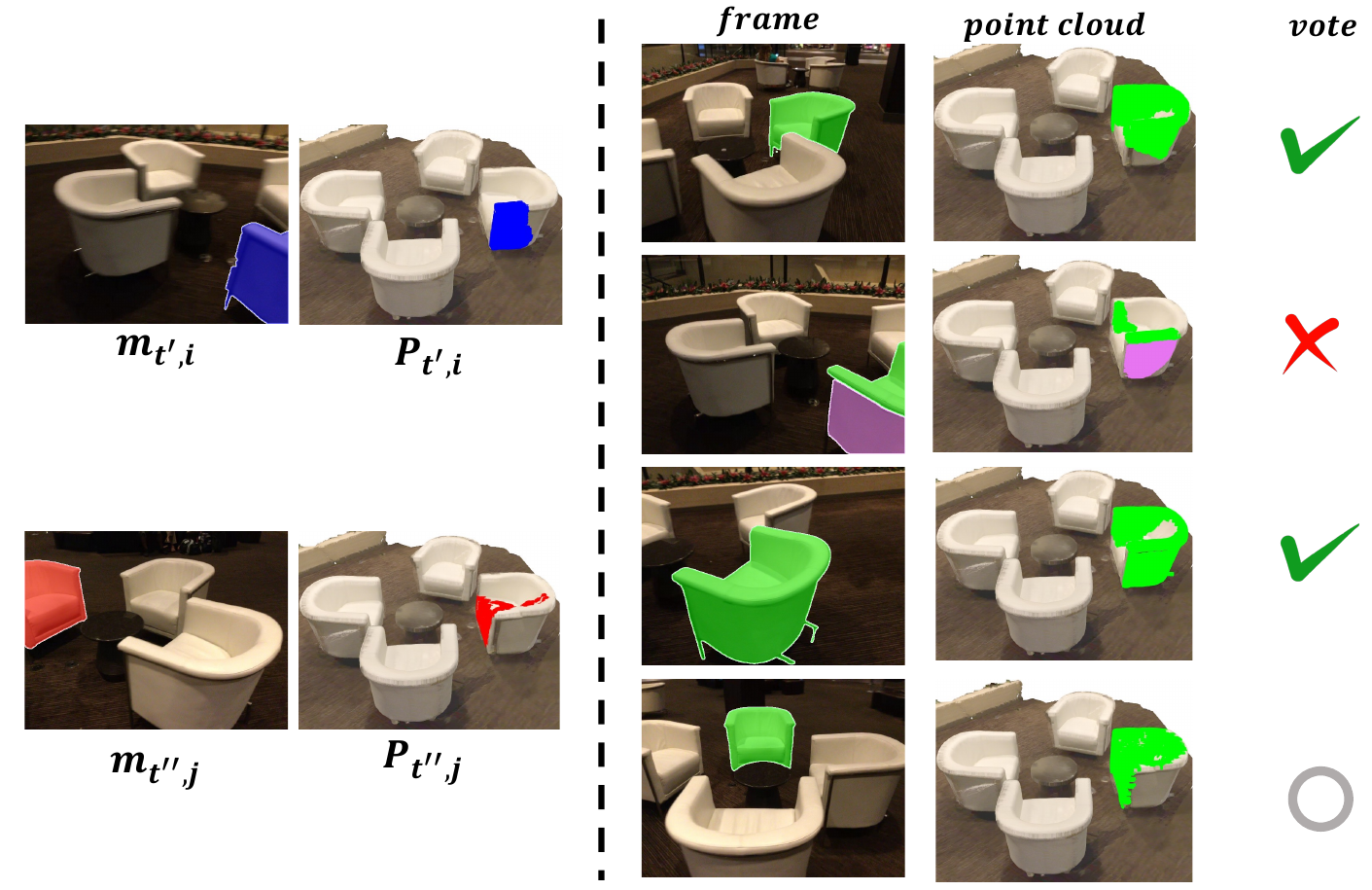}
\end{overpic}
\caption{
\textbf{View consensus rate.} Masks $m_{t',i}$ and $m_{t'',j}$ (side and frontal view of an armchair) are both visible in three frames, with two supporting them belonging to the same instance, resulting in a 2/3 consensus rate. Each mask is accompanied by its respective mask point cloud, displayed on the right. All point clouds are rendered under a consistent camera pose for clarity.
}

\label{fig:consensus}
\end{figure}

We then check whether an observer frame $t\in O$ supports the merging of these two masks. For an observer frame $t\in O$, if there exists a mask $m_{t,k}$ whose corresponding point cloud $P_{t,k}$ approximately contains both the point clouds $P_{t',i}^{t}$ of $m_{t',i}$ and $P_{t'',j}^{t}$ of $m_{t'',j}$, i.e., $P_{t',i}^{t} \sqsubset P_{t,k}$ and $P_{t'',j}^{t} \sqsubset P_{t,k}$, then this observer supports that the two masks are components of the same instance. The total number of supporters would be $n_{supporter}(m_{t',i}, m_{t'',j}) = \vert \{t \in O(m_{t',i}, m_{t'',j}) \mid \exists k, s.t.~P_{t',i}^t, P_{t'',j}^t \sqsubset P_{t,k}\}\vert$.
The proportion of supporters among all observers is subsequently defined as the \textbf{view consensus rate} $c$, as illustrated below:
\begin{equation}
\label{con_ratio_1}
    c(m_{t',i}, m_{t'',j}) = \frac{n_{supporter}(m_{t',i}, m_{t'',j})}{n(m_{t',i}, m_{t'',j})}
\end{equation}
 An illustration of thie view consensus rate can be found in Fig. \ref{fig:consensus}.

Employing the consensus rate as a criterion, we connect edges between mask pairs whose view consensus rates exceeding a predefined threshold $\tau_{rate}=0.9$. This procedure yields the set of edges $E$ as follows:
\begin{equation}
    E = \{ \{m_{t', i'}, m_{t'', i''}\} \mid c(m_{t',i'}, m_{t'',i''}) \geq \tau_{rate} \}
\end{equation}

Leveraging predictions across the entire sequence of images, our criterion shows enhanced robustness against over-segmentation errors compared to approaches that solely depend on local geometric overlap. Illustrated in Fig. \ref{fig:consensus}, the two masks exhibit low geometric overlap despite belonging to the same armchair. However, our approach identify a high consensus rate for them. This is attributed to the outstanding overall performance of modern mask predictors, which consistently segment this armchair comprehensively in most frames, encompassing both parts and thus yielding a high view consensus rate.

\subsubsection{Efficient Computation of View Consensus Rate}
\label{efficient_consensus}
Naively computing view consensus rates for all mask pairs can be untractable with  a time complexity of $\mathcal{O}(N^2T)$, where $N$ represents the total number of  masks, i.e., $N = \sum_{t} n_t$. To speed up, we initially calculate and store the intermediate result to eliminate redundant computations. 

Specifically, for each mask $m_{t',i}$, we first find $F(m_{t',i})$ and then identify all the \textit{masks that approximately contain it}, denoted as $M(m_{t',i}) = \{m_{t,k}\mid  t \in F(m_{t',i}) ~\text{and}~ P_{t',i}^{t} \sqsubset P_{t,k} \}$. With these intermediate results, the computation of equation \ref{con_ratio_1} can be simplified as,
\begin{equation}
\label{con_ratio_2}
    c(m_{t',i}, m_{t'',j}) = \frac{\vert M(m_{t',i}) \cap M(m_{t'',j}) \vert}{\vert F(m_{t',i}) \cap F(m_{t'',j}) \vert}
\end{equation}
In this way, all the operations in this expression have been simplified to simple set intersection operations involving only a few dozen elements, leading to a significant reduction in computational complexity. 

We now introduce the efficient computation of $M(m_{t',i})$. Initially, we examine the mask ID distribution of $P^t_{t', i}$ at frame $t$. If this distribution is concentrated, with more than $\tau_{contain}=0.8$ of elements equalling $k$, it indicates that $P^t_{t', i}$ primarily constitutes a part of the $k$-th instance at frame $t$. By definition, $P^t_{t', i} \sqsubset P_{t,k}$. The mask ID distribution is denoted as $d(m_{t',i}, t)$, and we elaborate on its efficient calculation through a space-time trade-off in the supplementary material.

\subsubsection{Under-Segment Mask Filtering}
\label{mask_filter}
We can also identify whether a mask is under-segmented based on the mask ID distribution $d(m_{t',i}, t)$. If $d(m_{t',i}, t)$ exhibits a very diverse distribution, it signifies that $P_{t',i}$ comprises multiple instances at frame $t$, making it highly likely that $m_{t', i}$ is an under-segmented mask. Assuming most 2D mask predictor outputs are correct, we ignore the alternative explanation that $m_{t',i}$ is accurate but the mask predictor over-segments this object consistently in other views.

Therefore, under-segmentation is marked by frequent distinction of $P_{t',i}$ into parts. We track the frequency of such occurrences (number of frames with diverse distributions in $d(m_{t',i}, t)$ / $|F(m_{t',i})|$). If this frequency exceeds $\tau_{filter}=0.2$, we classify the mask as under-segmented and filter it out. Specifically, we remove it from the mask graph. Additionally, to prevent this mask from erroneously inflating the consensus rate between two masks belonging to different instances, we also eliminate it from all $M(m_{t'', j})$ and remove $t'$ from $F(m_{t'',j})$.

\subsection{Iterative Graph Clustering}
\label{iterative}
Building upon the mask graph, we introduce an iterative graph clustering technique to merge masks and update the graph structure alternately. In the last iteration, each cluster denotes an instance.

When determining which masks to merge, we consider two strategies: 
1) merging each maximal clique (where a clique is a subset of the graph with an edge between every pair of nodes); 
2) merging each connected component (where a connected component is a subset of the graph  with a path between every pair of nodes). 
The first approach, though precise, tends to be overly stringent, often leading to insufficient merging and excessive over-segmentation. 
The second approach, more permissive, relies on the correctness of every pair-wise identified same-instance relationship, which can be less reliable when the number of observers $n$—the denominator of $c$—is low.

To balance these strategies, we modify the second approach to prioritize merging masks with a high number of observers first, postponing less reliable connections to later iterations.

\begin{figure}[t]
\centering
\begin{overpic}
[width=1\linewidth]
{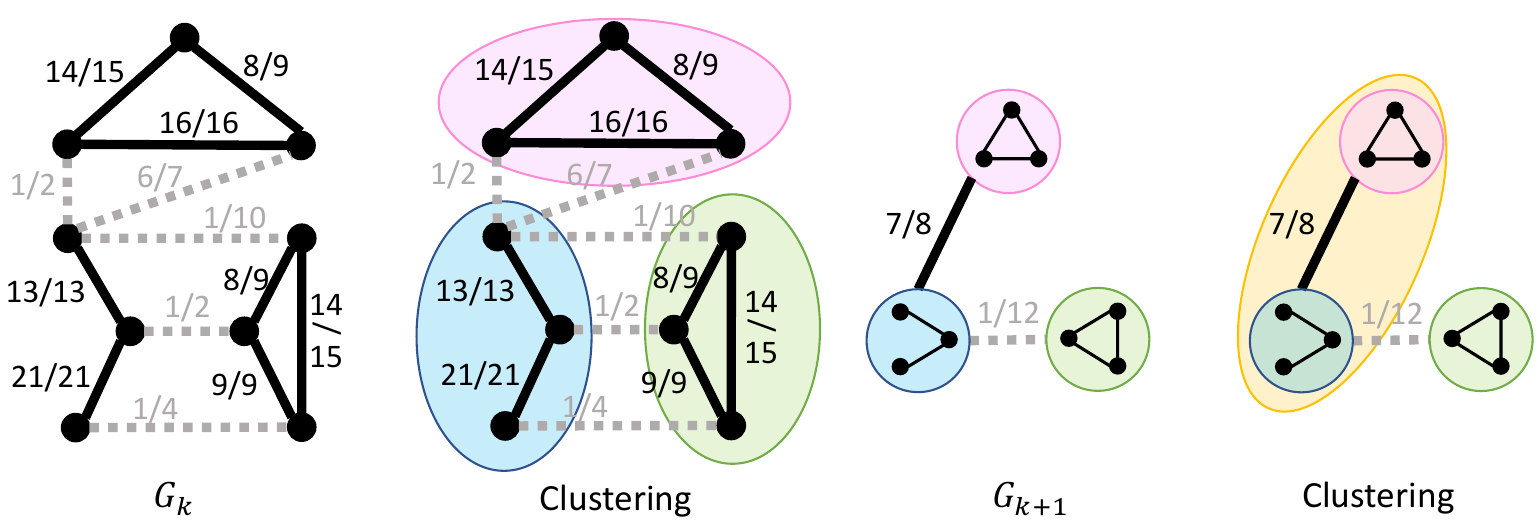}
\end{overpic}
\caption{
Illustration of iterative clustering. Node pairs with more observers are prioritized clustered ($G_k$). Then, view consensus of grouped masks is updated for the next clustering with more confident view consensus measurements. The text on the edge means $n_{support}/n$.}

\vspace{-10pt}
\label{fig:iterative}
\end{figure}

As illustrated in Fig.\ref{fig:iterative}, in each iteration $k$, we set an observer threshold $n_k$ and edges with $n < n_k$ are disconnected. We then identify connected components in the graph and merge them into new nodes. For a newly formed mask $m_{new}$ from a set of masks $\{m_{t_1, i_1}, m_{t_2, i_2}, \ldots, m_{t_s, i_s}\}$, its point cloud $P_{new}$ is the union of $\{P_{t_1, i_1}, P_{t_2, i_2}, \ldots, P_{t_s, i_s}\}$.

Subsequent to these node merging operations, updating edges requires recalculating the view consensus rate for the new mask in relation to others. Referring to equation \ref{con_ratio_2}, we calculate $F(m_{new})$ and $M(m_{new})$. While these two sets can be computed using the same technique as introduced in Sec.\ref{efficient_consensus}, we propose a method to accelerate this calculation while achieving comparable results through a straightforward approximation. Specifically, we approximate $F(m_{new})$ as $F(m_{t_1,i_1})\cup F(m_{t_2,i_2}) \ldots \cup F(m_{t_s, i_s})$ and $M(m_{new})$ as $M(m_{t_1,i_1})\cup M(m_{t_2,i_2}) \ldots \cup M(m_{t_s, i_s})$. This approximation is justified since masks merged due to high consensus rates often share containment by the same mask in frames where they both appear. The quantitative impact of this approximation is presented in Table \ref{table:clustering}.


After each iteration $k$, a new graph $G_{k+1}$ is formed. The observer threshold $n_k$ is adjusted downwards over several iterations to avoid neglecting smaller objects visible in fewer frames. We adopt a decreasing $n_k$ schedule, ranging from the top 5\%, 10\%, to 95\% of observer counts across all mask pairs. 

\subsection{Open-Vocabulary Feature Aggregation}
\label{open-voc}
After multiple iterations of clustering, we have obtained a conclusive list where each entry represents a 3D instance proposal. Simultaneously, we maintain a corresponding list of masks associated with each instance. This 2D-3D relationship allows us to directly select representative masks and fuse their semantic features to create an open-vocabulary feature for this instance. Following OpenMask3D\cite{takmaz2023openmask3d}, we first pick the top-5 masks that best cover the instance. Subsequently, we crop the original RGB image at multiple scales around each mask and input these image crops into CLIP\cite{radford2021learning} to extract open-vocabulary features. The final instance feature is derived from the average pooling result of these features.

\subsection{Implementation Details}

\label{implementation}
In order to obtain object-level masks, we use CropFormer \cite{qi2023high} as our 2D mask predictor. For open-vocabulary feature extraction, we use CLIP\cite{radford2021learning} ViT-H. To get mask point cloud $P_{t,i}$, we first back-project each mask to get the raw point cloud and then ball query the reconstructed point cloud with a radius equal to 3cm. We adopt the post-processing approach from OVIR-3D\cite{Lu2023OVIR3DO3} to refine the output 3D instances by using DBSCAN algorithm to separate disconnected point clusters into distinct instances.

\begin{table*}[htbp]
\caption{\textbf{Zero-shot 3D instance segmentation results on ScanNet++ and MatterPort3D.}  We report both semantic and class-agnostic performance. Our method outperform all baselines on all metrics significantly.}
\label{table:main}
\centering
\resizebox{1.6\columnwidth}{!}{
    \begin{tabular}{l|ccc|ccc|ccc|ccc}
    \toprule[2pt]
    \textbf{Model} & \multicolumn{6}{c|}{\textbf{ScanNet++}} & \multicolumn{6}{c}{\textbf{MatterPort3D}}\\ \hline
    & \multicolumn{3}{c|}{Class-agnostic} & \multicolumn{3}{c|}{Semantic}& \multicolumn{3}{c|}{Class-agnostic} & \multicolumn{3}{c}{Semantic}\\ \hline
          & $AP$ & $AP_{50}$ & $AP_{25}$ & $AP$ & $AP_{50}$ & $AP_{25}$ & $AP$ & $AP_{50}$ & $AP_{25}$ & $AP$ & $AP_{50}$ & $AP_{25}$ \\ \hline
          Mask3D& 22.8 & 33.3 & 45.7 
          & 3.6 & 5.1 & 6.7
          & 4.4 & 9.8 & 20.6
          & 2.6 & 4.7 & 7.0\\
          OpenMask3D 
          & 22.8 & 33.3 & 45.7
          & 2.0 & 2.7 & 3.4
          & 4.4 & 9.8 & 20.6
          & 4.6 & 8.5 & 13.0\\
         OVIR-3D
         & 19.4 & 34.1 & 46.5
         & 3.6 & 5.7 & 7.3
         & 5.9 & 13.9 & 24.6
         &  6.8 & 17.5 &  26.4 \\ 
         Ours 
         & \textbf{27.9} & \textbf{42.8} & \textbf{54.7}
         & \textbf{7.8} &  \textbf{10.7} & \textbf{12.1} 
         & \textbf{9.1} &  \textbf{19.5} & \textbf{35.3}
         & \textbf{11.1} & \textbf{21.1}  & \textbf{31.2}\\
    \bottomrule[2pt]
    \end{tabular}
}
\end{table*}

\section{Experiments}
In this section, we extensively evaluate our proposed method by comparing it with previous state-of-the-art methods on publicly available 3D instance segmentation benchmarks. 
The experimental setup is detailed in Section \ref{exp_setup}, and the statistics are comprehensively analyzed in Section \ref{quantitative}. 
Following that, we showcase the remarkable visual outcomes of our approach across a diverse range of complex scenes in Section \ref{qualitative}. The validation of all the components of our method is presented in Section \ref{ablation}.
\subsection{Experimental setup}
\label{exp_setup}

\textbf{Dataset}
ScanNet++\cite{yeshwanth2023scannet++} is a recently released high-quality benchmark that comprises 1554 classes with fine-grained annotation, making it an optimal choice for assessing open-vocabulary 3D instance segmentation. We also assess our method on two widely-used benchmarks: ScanNet200\cite{dai2017scannet, rozenberszki2022language}, which focuses on room-level evaluations, and MatterPort3D\cite{Matterport3D}, designed for building-level evaluations with sparser viewpoints. We utilize the validation sets of ScanNet++ and ScanNet, along with the testing set of MatterPort3D.


\noindent\textbf{Baselines}
We select the recent SOTA methods on both supervised closed-set 3D instance segmentation and open-vocabulary 3D instance segmentation.
Mask3D \cite{schult2022mask3d} stands out as a state-of-the-art method which requires supervised training on ScanNet200. OpenMask3D \cite{takmaz2023openmask3d} leverages supervised mask proposals from Mask3D and employs CLIP for open-vocabulary semantics aggregation. Different from our setting, both of them rely on supervised mask. OVIR-3D \cite{Lu2023OVIR3DO3} utilize both zero-shot masks and semantics, merging zero-shot 2D masks with large geometric and semantic overlap and using K-Means to choose the most representative features from the per-frame semantic feature.

\noindent\textbf{Metrics}
We report the standard Average Precision (AP) at 25\% and 50\% IoU and the mean of AP from 50\% to 95\% at 5\% intervals. In addition to the conventional semantic instance segmentation setting, we also test in a class-agnostic setting, disregarding semantic labels and solely assessing mask quality. This setting offers a precise assessment of the zero-shot mask prediction capability.

\begin{table}[htbp]
\caption{\textbf{3D instance segmentation results on ScanNet200.}  Mask3D and OpenMask3D both require supervised (sup.) training on ScanNet200. In fully zero shot (z.s.) setting, our method surpass OVIR-3D by a large margin on all metrics.   }
\label{table:scannet200}
\centering
\resizebox{1\columnwidth}{!}{
    \begin{tabular}{lcccccc}
    \toprule[2pt]
     \multirow{2}*{\textbf{Model}} &   \multicolumn{3}{c}{\textbf{Class-agnostic}} & \multicolumn{3}{c}{\textbf{Semantic}} \\ 
     \cline{2-7}
      ~ & AP & $AP_{50}$ & $AP_{25}$ & AP & $AP_{50}$ & $AP_{25}$ \\ \hline
      \textit{sup. mask + sup. semantic}  \\
      
      Mask3D&  39.7 & 53.6& 62.5 & 26.9 & 36.2 & 41.4\\ \hline
      \textit{sup. mask + z.s. semantic}  \\
      
      OpenMask3D & 39.7 & 53.6 & 62.5 & 15.1& 19.6 & 22.6\\ \hline
      \textit{z.s. mask + z.s. semantic}  \\
     OVIR-3D& 14.4 & 27.5 & 38.8 & 9.3& 18.7& 25.0\\ 
     Ours& \textbf{19.2} & \textbf{36.6} & \textbf{51.7}& \textbf{12.0} & \textbf{23.3} & \textbf{30.1}\\
    \bottomrule[2pt]
    \end{tabular}
}
\end{table}

\subsection{Quantitative Comparison.}
\label{quantitative}

\noindent{\textbf{ScanNet++ and MatterPort3D.}} We directly test all methods on ScanNet++ and MatterPort3D in a zero-shot manner. As shown in Table \ref{table:main}, our method outperforms all baselines by a large margin. In comparison to OVIR-3D, the most akin work to ours, we achieve +4.1\% and +4.7\% AP on ScanNet++ semantic and class-agnostic setting, respectively.  Similarly, we demonstrate +3.0\% and +2.0\% AP on MatterPort3D in the same settings, validating our globally optimal association design.

OpenMask3D shares Mask3D's mask predictor, rendering their performance identical in the class-agnostic setting. We observe that this mask predictor, trained on ScanNet200, has limited generalizability. While it shows impressive results within the confines of ScanNet200, its performance suffers significantly when evaluated on new benchmarks, as demonstrated in Table \ref{table:main} and Fig. \ref{fig:gallery}. Additionally, it exhibits sensitivity to point distribution patterns. For instance, in the class-agnostic setting of ScanNet++, its AP is a mere 13.6\% when using the raw point cloud as input. Interestingly, a simple preprocessing step such as uniform sampling significantly boosts performance to 22.8\%.

\noindent{\textbf{ScanNet200.}} As the mask predictor and semantic head of Mask3D are trained specifically on ScanNet200, we classify methods according to their train-test settings. In comparison to the fully zero-shot method, OVIR-3D, our approach exhibits a significant performance advantage, surpassing it by +5.3\% in average precision (AP), +8.9\% in $AP_{50}$, and +12.6\% in $AP_{25}$ in the class-agnostic setting. This further underscores the effectiveness of our proposed globally optimal merging mechanism. Moreover, our method even outperforms OpenMask3D, which relies on a supervised mask predictor, by a substantial margin of +3.7\% in $AP_{50}$ and +7.5\% in $AP_{25}$.

\subsection{Ablation Studies} 
\label{ablation}
In Table \ref{table:Ablation}, we analyze key components of our method on ScanNet200—under-segment mask filtering and iterative clustering. Starting with a baseline using view consensus rate, we merge masks within connected components. This simple approach matches OVIR-3D baseline performance. Upon adding under-segment mask filtering and iterative clustering, performance steadily rises from 10.0\% $AP$ to 11.7\% $AP$, reaching peak performance when both modules are combined.

\begin{table}[htbp]
\caption{\textbf{Ablation study on under-segment mask filtering and iterative clustering on ScanNet200.}}
\label{table:Ablation}
\centering
    \resizebox{0.9\columnwidth}{!}{
        \begin{tabular}{cc|ccc}
        
            \toprule[2pt]
             under. filtering & iter. clustering & $AP$ & $AP_{50}$ & $AP_{25}$ \\ \hline
              \ding{56} & \ding{56} & 10.0 & 19.1 & 24.2 \\ \hline
            \ding{52} & \ding{56} & 11.0 & 21.2 & 27.5\\ \hline
            \ding{56} & \ding{52} & 11.7 & 22.3 & 29.2 \\ \hline
            \ding{52} & \ding{52} & \textbf{12.0} & \textbf{23.3} & \textbf{30.1}\\
            \bottomrule[2pt]
        \end{tabular}
    }
\end{table}
\vspace{-3mm}

We compare various clustering algorithms, including clustering cliques or connected components as discussed in Section \ref{iterative}, and also clustering a relaxation of cliques using the Highly Connected Sub-graphs (HCS) algorithm~\cite{hartuv2000clustering}. We also show the impact of the approximation introduced in Section \ref{iterative}. As shown in Table \ref{table:clustering}, our proposed iterative clustering method outperforms all other trials. Comprehensive statistics are available in the supplementary material.

\vspace{-2mm}
\begin{table}[htbp]
\caption{\textbf{Ablation study on clustering methods.}}
\label{table:clustering}
\centering
\resizebox{0.8\columnwidth}{!}{
    \begin{tabular}{l|ccc}
    \toprule[2pt]
    Clustering Algorithm & AP & $AP_{50}$ & $AP_{25}$ \\ \hline
     Connected component & 11.0 & 21.2 & 27.5\\
     Clique & 11.3 & 22.0 & 29.4\\
     Quasi-Clique (HCS) & 11.9 & 22.9 & 29.7\\
     Ours w/o approximation & 11.8 & 23.1 & \textbf{30.4}\\
     Ours& \textbf{12.0} & \textbf{23.3} & 30.1\\
    \bottomrule[2pt]
    \end{tabular}
}
\end{table}

We conducted additional evaluations to assess the robustness of our algorithm to variations in hyperparameters. For the mask visibility threshold $\tau_{vis}$ ranging from 0.6 to 0.8, the under-segment mask filtering threshold $\tau_{filter}$ ranging from 0.2 to 0.4, the consensus rate threshold $\tau_{rate}$ ranging from 0.8 to 1 and the approximate containment threshold $\tau_{contain}$ ranging from 0.7 to 0.9, our method consistently demonstrates satisfying performance.

\begin{table}[t]
\caption{\textbf{Ablation study on Hyperparameters on ScanNet200.}}
\label{table:hyper}
    \resizebox{\columnwidth}{!}{
        \begin{tabular}{l|ccc}
            \toprule[2pt]
            & $AP$ & $AP_{50}$ & $AP_{25}$ \\ \hline
            $\tau_{vis} (0.6-0.8)$ & 11.9 $\pm$ 0.06 & 23.2 $\pm$ 0.09 & 30.1 $\pm$ 0.07 \\ \hline
            $\tau_{filter} (0.2-0.4)$ & 11.9 $\pm$ 0.05 & 23.3 $\pm$ 0.19 & 30.0 $\pm$ 0.18 \\ \hline
            $\tau_{rate} (0.8-1)$ & 11.8 $\pm$ 0.20 & 22.7 $\pm$ 0.52 & 28.9 $\pm$ 0.83 \\ \hline
            $\tau_{contain} (0.7-0.9)$ & 11.9$\pm$0.10 & 23.3$\pm$ 0.22 & 30.3$\pm$ 0.20 \\
            \bottomrule[2pt]
        \end{tabular}
    }
\end{table}

\subsection{Qualitative Results.} 
\label{qualitative}
\begin{figure*}[t]
\centering
\begin{overpic}
[width=1\linewidth]
{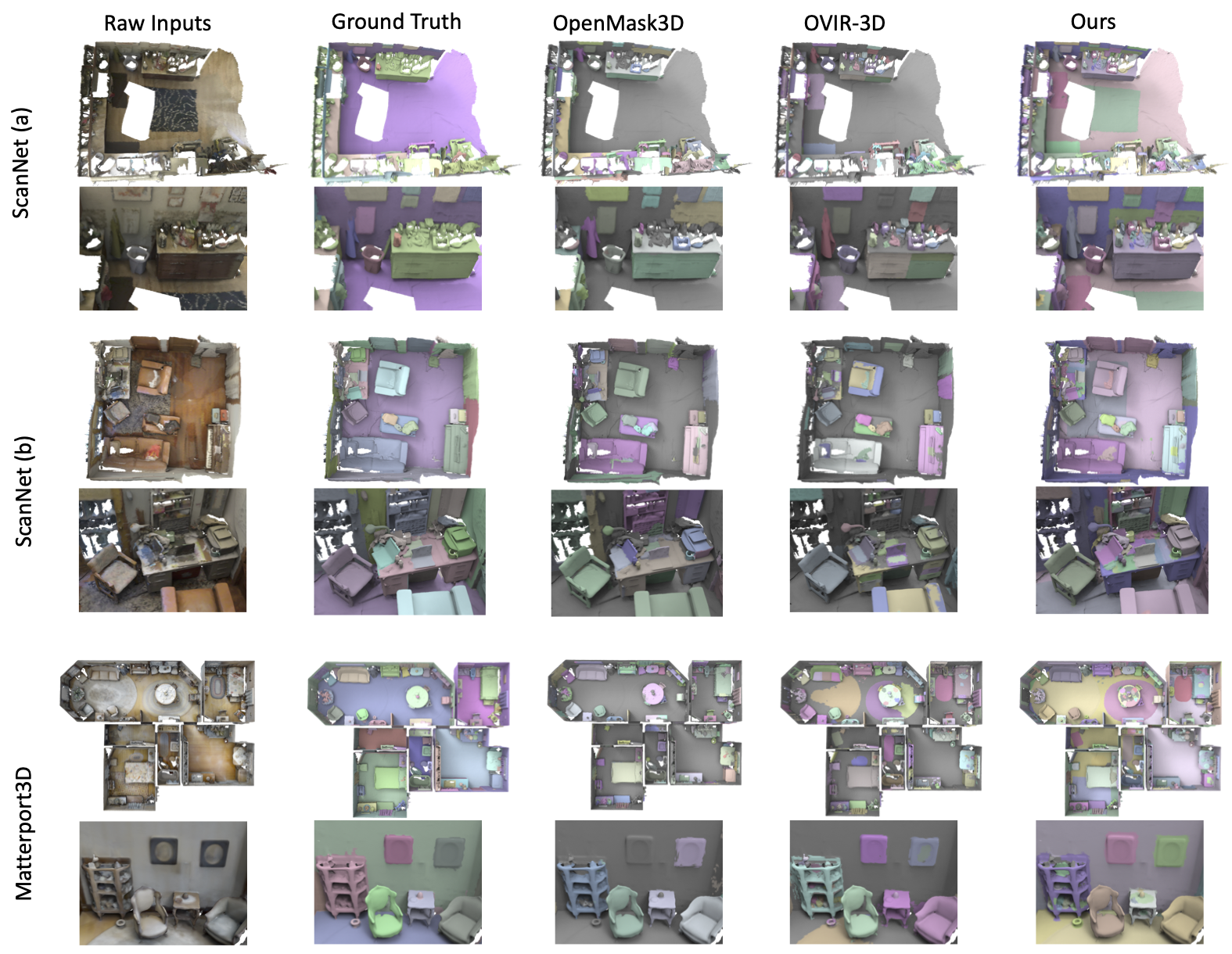}
\end{overpic}
\caption{
Comparison of 3D zero-shot segmentation performance. We compare our methods with OpenMask3D~\cite{takmaz2023openmask3d} and OVIR-3D~\cite{Lu2023OVIR3DO3} on ScanNet~\cite{dai2017scannet} and Matterport3D~\cite{Matterport3D}.}
\label{fig:gallery}
\end{figure*}

\begin{figure}[htbp]
\centering
\includegraphics[width=0.9\columnwidth]{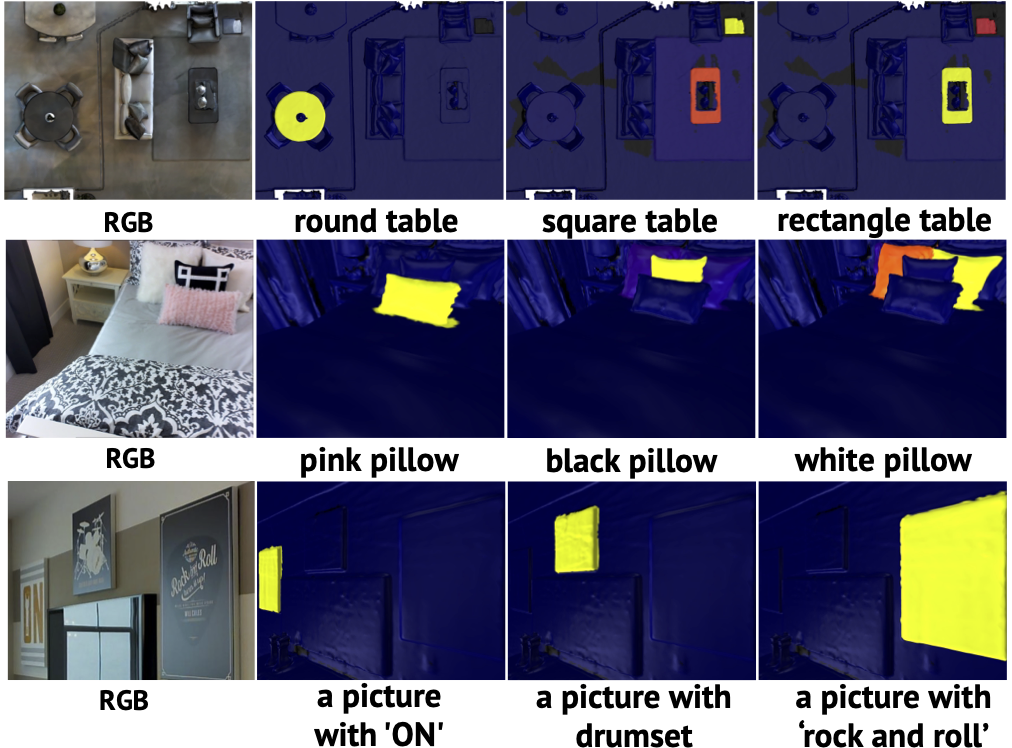}
\caption{
\textbf{Open-vocabulary queries of different shapes, colors and contents.} 
}
\label{fig:open-voc-gallary}
\end{figure}

In Fig. \ref{fig:open-voc-gallary}, we present the similarity heatmaps for a wide range of open-vocabulary queries, showcasing the remarkable capabilities of our open-vocabulary segmentation system. Additionally, in Fig. \ref{fig:gallery}, we offer a visual comparison of our algorithm against all baseline methods. Our method shows excellent ability to segment small objects, \textit{e.g.}, items on the counter in ScanNet a), boxes on the shelf in ScanNet b). These small objects are simply labeled as part of its containers in the ground truth, which cause the AP at higher IoU threshold of our method drops severely.

Compared to OVIR-3D, our method has two main advantages: i) OVIR-3D can't merges masks that have low geometric overlap but correspond to a same object well. For example, in ScanNet (b), items on the coffee table split the table point cloud into two pieces, making OVIR-3D fail to merge these two parts together. So do the sofa chair in ScanNet b) and the right rug in the MatterPort3D example. In the contrary, our method merges these objects well based on view consensus as explained in Section \ref{view_consensus}. ii) The strict filtering process in OVIR-3D falsely filter out many objects, \textit{e.g.}, counter and pictures in ScanNet a) while our method only conservatively filter out under-segment masks.

\subsection{Limitations}
While our approach demonstrates remarkable performance, it is important to acknowledge two notable limitations. Firstly, this work assumes near-perfect 2D segmentation and 2D-3D correspondence, which may not always be the case in certain applications. Presently, we only generate object-level masks, whereas real-world applications may necessitate multi-level masks spanning from parts and objects to clusters.
\section{Conclusion}
In this work, we propose a view consensus based mask graph clustering algorithm for open-vocabulary 3D instance segmentation. Specifically, our method constructs a global mask graph and leverages the view consensus to cluster the masks belonging to the same 3D instances. Besides, the mask clustering guided the clustering of the open-vocabulary features for text queries. The results demonstrate that our method achieves SOTA performance on zero-shot mask prediction and open-vocabulary understating. In the future, we would like to investigate the application of the proposed method on robotic tasks, such as open-vocabulary object navigation.

\section{Acknowledgements}
This work was supported in part by National Key R\&D Program of China 2022ZD0160801.

\newpage
{
    \small
    \bibliographystyle{ieeenat_fullname}
    \bibliography{main}

\begin{thebibliography}{53}
\providecommand{\natexlab}[1]{#1}
\providecommand{\url}[1]{\texttt{#1}}
\expandafter\ifx\csname urlstyle\endcsname\relax
  \providecommand{\doi}[1]{doi: #1}\else
  \providecommand{\doi}{doi: \begingroup \urlstyle{rm}\Url}\fi

\bibitem[Chang et~al.(2017)Chang, Dai, Funkhouser, Halber, Niessner, Savva, Song, Zeng, and Zhang]{Matterport3D}
Angel Chang, Angela Dai, Thomas Funkhouser, Maciej Halber, Matthias Niessner, Manolis Savva, Shuran Song, Andy Zeng, and Yinda Zhang.
\newblock Matterport3d: Learning from rgb-d data in indoor environments.
\newblock \emph{International Conference on 3D Vision (3DV)}, 2017.

\bibitem[Cheng et~al.(2022)Cheng, Misra, Schwing, Kirillov, and Girdhar]{cheng2022masked}
Bowen Cheng, Ishan Misra, Alexander~G Schwing, Alexander Kirillov, and Rohit Girdhar.
\newblock Masked-attention mask transformer for universal image segmentation.
\newblock In \emph{Proceedings of the IEEE/CVF conference on computer vision and pattern recognition}, pages 1290--1299, 2022.

\bibitem[Choy et~al.(2019)Choy, Gwak, and Savarese]{choy20194d}
Christopher Choy, JunYoung Gwak, and Silvio Savarese.
\newblock 4d spatio-temporal convnets: Minkowski convolutional neural networks.
\newblock In \emph{Proceedings of the IEEE/CVF conference on computer vision and pattern recognition}, pages 3075--3084, 2019.

\bibitem[Dai et~al.(2017)Dai, Chang, Savva, Halber, Funkhouser, and Nie{\ss}ner]{dai2017scannet}
Angela Dai, Angel~X Chang, Manolis Savva, Maciej Halber, Thomas Funkhouser, and Matthias Nie{\ss}ner.
\newblock Scannet: Richly-annotated 3d reconstructions of indoor scenes.
\newblock In \emph{Proceedings of the IEEE conference on computer vision and pattern recognition}, pages 5828--5839, 2017.

\bibitem[Ester et~al.(1996)Ester, Kriegel, Sander, Xu, et~al.]{ester1996density}
Martin Ester, Hans-Peter Kriegel, J{\"o}rg Sander, Xiaowei Xu, et~al.
\newblock A density-based algorithm for discovering clusters in large spatial databases with noise.
\newblock In \emph{kdd}, pages 226--231, 1996.

\bibitem[Felzenszwalb and Huttenlocher(2004)]{felzenszwalb2004efficient}
Pedro~F Felzenszwalb and Daniel~P Huttenlocher.
\newblock Efficient graph-based image segmentation.
\newblock \emph{International journal of computer vision}, 59:\penalty0 167--181, 2004.

\bibitem[Ghiasi et~al.(2021)Ghiasi, Gu, Cui, and Lin]{Ghiasi2021ScalingOI}
Golnaz Ghiasi, Xiuye Gu, Yin Cui, and Tsung-Yi Lin.
\newblock Scaling open-vocabulary image segmentation with image-level labels.
\newblock In \emph{European Conference on Computer Vision}, 2021.

\bibitem[Gu et~al.(2023)Gu, Kuwajerwala, Morin, Jatavallabhula, Sen, Agarwal, Rivera, Paul, Ellis, Chellappa, Gan, de~Melo, Tenenbaum, Torralba, Shkurti, and Paull]{Gu2023ConceptGraphsO3}
Qiao Gu, Ali Kuwajerwala, Sacha Morin, Krishna~Murthy Jatavallabhula, Bipasha Sen, Aditya Agarwal, Corban Rivera, William Paul, Kirsty Ellis, Ramalingam Chellappa, Chuang Gan, Celso~Miguel de Melo, Joshua~B. Tenenbaum, Antonio Torralba, Florian Shkurti, and Liam Paull.
\newblock Conceptgraphs: Open-vocabulary 3d scene graphs for perception and planning.
\newblock \emph{ArXiv}, abs/2309.16650, 2023.

\bibitem[Han et~al.(2020)Han, Zheng, Xu, and Fang]{han2020occuseg}
Lei Han, Tian Zheng, Lan Xu, and Lu Fang.
\newblock Occuseg: Occupancy-aware 3d instance segmentation.
\newblock In \emph{Proceedings of the IEEE/CVF conference on computer vision and pattern recognition}, pages 2940--2949, 2020.

\bibitem[Hartuv and Shamir(2000)]{hartuv2000clustering}
Erez Hartuv and Ron Shamir.
\newblock A clustering algorithm based on graph connectivity.
\newblock \emph{Information processing letters}, 76\penalty0 (4-6):\penalty0 175--181, 2000.

\bibitem[He et~al.(2023)He, Ding, and Jiang]{He2023SemanticPromotedDA}
Shuting He, Henghui Ding, and Wei Jiang.
\newblock Semantic-promoted debiasing and background disambiguation for zero-shot instance segmentation.
\newblock \emph{2023 IEEE/CVF Conference on Computer Vision and Pattern Recognition (CVPR)}, pages 19498--19507, 2023.

\bibitem[Hou et~al.(2019)Hou, Dai, and Nie{\ss}ner]{hou20193d}
Ji Hou, Angela Dai, and Matthias Nie{\ss}ner.
\newblock 3d-sis: 3d semantic instance segmentation of rgb-d scans.
\newblock In \emph{Proceedings of the IEEE/CVF conference on computer vision and pattern recognition}, pages 4421--4430, 2019.

\bibitem[Hu et~al.(2021{\natexlab{a}})Hu, Zhao, Jiang, Jia, and Wong]{hu2021bidirectional}
Wenbo Hu, Hengshuang Zhao, Li Jiang, Jiaya Jia, and Tien-Tsin Wong.
\newblock Bidirectional projection network for cross dimension scene understanding.
\newblock In \emph{Proceedings of the IEEE/CVF Conference on Computer Vision and Pattern Recognition}, pages 14373--14382, 2021{\natexlab{a}}.

\bibitem[Hu et~al.(2021{\natexlab{b}})Hu, Bai, Shang, Zhang, Dong, Wang, Sun, Fu, and Tai]{hu2021vmnet}
Zeyu Hu, Xuyang Bai, Jiaxiang Shang, Runze Zhang, Jiayu Dong, Xin Wang, Guangyuan Sun, Hongbo Fu, and Chiew-Lan Tai.
\newblock Vmnet: Voxel-mesh network for geodesic-aware 3d semantic segmentation.
\newblock In \emph{Proceedings of the IEEE/CVF International Conference on Computer Vision}, pages 15488--15498, 2021{\natexlab{b}}.

\bibitem[Huang et~al.(2023{\natexlab{a}})Huang, Mees, Zeng, and Burgard]{huang2023visual}
Chenguang Huang, Oier Mees, Andy Zeng, and Wolfram Burgard.
\newblock Visual language maps for robot navigation.
\newblock In \emph{2023 IEEE International Conference on Robotics and Automation (ICRA)}, pages 10608--10615. IEEE, 2023{\natexlab{a}}.

\bibitem[Huang et~al.(2019)Huang, Zhang, Yi, Funkhouser, Nie{\ss}ner, and Guibas]{huang2019texturenet}
Jingwei Huang, Haotian Zhang, Li Yi, Thomas Funkhouser, Matthias Nie{\ss}ner, and Leonidas~J Guibas.
\newblock Texturenet: Consistent local parametrizations for learning from high-resolution signals on meshes.
\newblock In \emph{Proceedings of the IEEE/CVF Conference on Computer Vision and Pattern Recognition}, pages 4440--4449, 2019.

\bibitem[Huang et~al.(2021)Huang, Ma, Mu, Fu, and Hu]{huang2021supervoxel}
Shi-Sheng Huang, Ze-Yu Ma, Tai-Jiang Mu, Hongbo Fu, and Shi-Min Hu.
\newblock Supervoxel convolution for online 3d semantic segmentation.
\newblock \emph{ACM Transactions on Graphics (TOG)}, 40\penalty0 (3):\penalty0 1--15, 2021.

\bibitem[Huang et~al.(2023{\natexlab{b}})Huang, Wu, Chen, Zhao, Zhu, and Lasenby]{Huang2023OpenIns3DSA}
Zhening Huang, Xiaoyang Wu, Xi Chen, Hengshuang Zhao, Lei Zhu, and Joan Lasenby.
\newblock Openins3d: Snap and lookup for 3d open-vocabulary instance segmentation.
\newblock \emph{ArXiv}, abs/2309.00616, 2023{\natexlab{b}}.

\bibitem[Huang et~al.(2023{\natexlab{c}})Huang, Wu, Chen, Zhao, Zhu, and Lasenby]{huang2023openins3d}
Zhening Huang, Xiaoyang Wu, Xi Chen, Hengshuang Zhao, Lei Zhu, and Joan Lasenby.
\newblock Openins3d: Snap and lookup for 3d open-vocabulary instance segmentation.
\newblock \emph{arXiv preprint arXiv:2309.00616}, 2023{\natexlab{c}}.

\bibitem[Huynh et~al.(2021)Huynh, Kuen, nan Lin, Gu, and Elhamifar]{Huynh2021OpenVocabularyIS}
Dat~T. Huynh, Jason Kuen, Zhe nan Lin, Jiuxiang Gu, and Ehsan Elhamifar.
\newblock Open-vocabulary instance segmentation via robust cross-modal pseudo-labeling.
\newblock \emph{2022 IEEE/CVF Conference on Computer Vision and Pattern Recognition (CVPR)}, pages 7010--7021, 2021.

\bibitem[Kaul et~al.(2023)Kaul, Xie, and Zisserman]{kaul2023multi}
Prannay Kaul, Weidi Xie, and Andrew Zisserman.
\newblock Multi-modal classifiers for open-vocabulary object detection.
\newblock In \emph{International Conference on Machine Learning}, pages 15946--15969. PMLR, 2023.

\bibitem[Kerr et~al.(2023)Kerr, Kim, Goldberg, Kanazawa, and Tancik]{kerr2023lerf}
Justin Kerr, Chung~Min Kim, Ken Goldberg, Angjoo Kanazawa, and Matthew Tancik.
\newblock Lerf: Language embedded radiance fields.
\newblock In \emph{Proceedings of the IEEE/CVF International Conference on Computer Vision}, pages 19729--19739, 2023.

\bibitem[Kim et~al.(2023)Kim, Angelova, and Kuo]{kim2023region}
Dahun Kim, Anelia Angelova, and Weicheng Kuo.
\newblock Region-aware pretraining for open-vocabulary object detection with vision transformers.
\newblock In \emph{Proceedings of the IEEE/CVF Conference on Computer Vision and Pattern Recognition}, pages 11144--11154, 2023.

\bibitem[Kirillov et~al.(2023)Kirillov, Mintun, Ravi, Mao, Rolland, Gustafson, Xiao, Whitehead, Berg, Lo, et~al.]{kirillov2023segment}
Alexander Kirillov, Eric Mintun, Nikhila Ravi, Hanzi Mao, Chloe Rolland, Laura Gustafson, Tete Xiao, Spencer Whitehead, Alexander~C Berg, Wan-Yen Lo, et~al.
\newblock Segment anything.
\newblock In \emph{Proceedings of the IEEE/CVF International Conference on Computer Vision}, pages 4015--4026, 2023.

\bibitem[Li et~al.(2022{\natexlab{a}})Li, Weinberger, Belongie, Koltun, and Ranftl]{Li2022LanguagedrivenSS}
Boyi Li, Kilian~Q Weinberger, Serge Belongie, Vladlen Koltun, and Rene Ranftl.
\newblock Language-driven semantic segmentation.
\newblock In \emph{International Conference on Learning Representations}, 2022{\natexlab{a}}.

\bibitem[Li et~al.(2022{\natexlab{b}})Li, He, Wen, Gao, Cheng, and Zhang]{li2022panoptic}
Jinke Li, Xiao He, Yang Wen, Yuan Gao, Xiaoqiang Cheng, and Dan Zhang.
\newblock Panoptic-phnet: Towards real-time and high-precision lidar panoptic segmentation via clustering pseudo heatmap.
\newblock In \emph{Proceedings of the IEEE/CVF Conference on Computer Vision and Pattern Recognition}, pages 11809--11818, 2022{\natexlab{b}}.

\bibitem[Liu et~al.(2022)Liu, Zheng, Lin, Ni, and Fang]{liu2022ins}
Leyao Liu, Tian Zheng, Yun-Jou Lin, Kai Ni, and Lu Fang.
\newblock Ins-conv: Incremental sparse convolution for online 3d segmentation.
\newblock In \emph{Proceedings of the IEEE/CVF Conference on Computer Vision and Pattern Recognition}, pages 18975--18984, 2022.

\bibitem[Lu et~al.(2023)Lu, Chang, Jing, Boularias, and Bekris]{Lu2023OVIR3DO3}
Shiyang Lu, Haonan Chang, Eric~Pu Jing, Abdeslam Boularias, and Kostas Bekris.
\newblock Ovir-3d: Open-vocabulary 3d instance retrieval without training on 3d data.
\newblock In \emph{Conference on Robot Learning}, pages 1610--1620. PMLR, 2023.

\bibitem[Narita et~al.(2019)Narita, Seno, Ishikawa, and Kaji]{narita2019panopticfusion}
Gaku Narita, Takashi Seno, Tomoya Ishikawa, and Yohsuke Kaji.
\newblock Panopticfusion: Online volumetric semantic mapping at the level of stuff and things.
\newblock In \emph{2019 IEEE/RSJ International Conference on Intelligent Robots and Systems (IROS)}, pages 4205--4212. IEEE, 2019.

\bibitem[Nguyen et~al.(2023)Nguyen, Ngo, Gan, Kalogerakis, Tran, Pham, and Nguyen]{nguyen2023open3dis}
Phuc~DA Nguyen, Tuan~Duc Ngo, Chuang Gan, Evangelos Kalogerakis, Anh Tran, Cuong Pham, and Khoi Nguyen.
\newblock Open3dis: Open-vocabulary 3d instance segmentation with 2d mask guidance.
\newblock \emph{arXiv preprint arXiv:2312.10671}, 2023.

\bibitem[Peng et~al.(2023)Peng, Genova, Jiang, Tagliasacchi, Pollefeys, Funkhouser, et~al.]{peng2023openscene}
Songyou Peng, Kyle Genova, Chiyu Jiang, Andrea Tagliasacchi, Marc Pollefeys, Thomas Funkhouser, et~al.
\newblock Openscene: 3d scene understanding with open vocabularies.
\newblock In \emph{Proceedings of the IEEE/CVF Conference on Computer Vision and Pattern Recognition}, pages 815--824, 2023.

\bibitem[Qi et~al.(2023)Qi, Kuen, Shen, Gu, Li, Guo, Jia, Lin, and Yang]{qi2023high}
Lu Qi, Jason Kuen, Tiancheng Shen, Jiuxiang Gu, Wenbo Li, Weidong Guo, Jiaya Jia, Zhe Lin, and Ming-Hsuan Yang.
\newblock High quality entity segmentation.
\newblock In \emph{Proceedings of the IEEE/CVF International Conference on Computer Vision}, pages 4047--4056, 2023.

\bibitem[Radford et~al.(2021)Radford, Kim, Hallacy, Ramesh, Goh, Agarwal, Sastry, Askell, Mishkin, Clark, et~al.]{radford2021learning}
Alec Radford, Jong~Wook Kim, Chris Hallacy, Aditya Ramesh, Gabriel Goh, Sandhini Agarwal, Girish Sastry, Amanda Askell, Pamela Mishkin, Jack Clark, et~al.
\newblock Learning transferable visual models from natural language supervision.
\newblock In \emph{International conference on machine learning}, pages 8748--8763. PMLR, 2021.

\bibitem[Robert et~al.(2022)Robert, Vallet, and Landrieu]{robert2022learning}
Damien Robert, Bruno Vallet, and Loic Landrieu.
\newblock Learning multi-view aggregation in the wild for large-scale 3d semantic segmentation.
\newblock In \emph{Proceedings of the IEEE/CVF Conference on Computer Vision and Pattern Recognition}, pages 5575--5584, 2022.

\bibitem[Rozenberszki et~al.(2022)Rozenberszki, Litany, and Dai]{rozenberszki2022language}
David Rozenberszki, Or Litany, and Angela Dai.
\newblock Language-grounded indoor 3d semantic segmentation in the wild.
\newblock In \emph{European Conference on Computer Vision}, pages 125--141. Springer, 2022.

\bibitem[Schult et~al.(2023)Schult, Engelmann, Hermans, Litany, Tang, and Leibe]{schult2022mask3d}
Jonas Schult, Francis Engelmann, Alexander Hermans, Or Litany, Siyu Tang, and Bastian Leibe.
\newblock Mask3d: Mask transformer for 3d semantic instance segmentation.
\newblock In \emph{2023 IEEE International Conference on Robotics and Automation (ICRA)}, pages 8216--8223. IEEE, 2023.

\bibitem[Takmaz et~al.(2023)Takmaz, Fedele, Sumner, Pollefeys, Tombari, and Engelmann]{takmaz2023openmask3d}
Ay{\c{c}}a Takmaz, Elisabetta Fedele, Robert~W. Sumner, Marc Pollefeys, Federico Tombari, and Francis Engelmann.
\newblock {OpenMask3D: Open-Vocabulary 3D Instance Segmentation}.
\newblock In \emph{Advances in Neural Information Processing Systems (NeurIPS)}, 2023.

\bibitem[Triggs et~al.(2000)Triggs, McLauchlan, Hartley, and Fitzgibbon]{triggs2000bundle}
Bill Triggs, Philip~F McLauchlan, Richard~I Hartley, and Andrew~W Fitzgibbon.
\newblock Bundle adjustment—a modern synthesis.
\newblock In \emph{Vision Algorithms: Theory and Practice: International Workshop on Vision Algorithms Corfu, Greece, September 21--22, 1999 Proceedings}, pages 298--372. Springer, 2000.

\bibitem[Vibashan et~al.(2023)Vibashan, Yu, Xing, Qin, Gao, Niebles, Patel, and Xu]{Vibashan2023MaskFreeOO}
VS Vibashan, Ning Yu, Chen Xing, Can Qin, Mingfei Gao, Juan~Carlos Niebles, Vishal~M. Patel, and Ran Xu.
\newblock Mask-free ovis: Open-vocabulary instance segmentation without manual mask annotations.
\newblock \emph{2023 IEEE/CVF Conference on Computer Vision and Pattern Recognition (CVPR)}, pages 23539--23549, 2023.

\bibitem[Vora et~al.(2021)Vora, Radwan, Greff, Meyer, Genova, Sajjadi, Pot, Tagliasacchi, and Duckworth]{vora2021nesf}
Suhani Vora, Noha Radwan, Klaus Greff, Henning Meyer, Kyle Genova, Mehdi~SM Sajjadi, Etienne Pot, Andrea Tagliasacchi, and Daniel Duckworth.
\newblock Nesf: Neural semantic fields for generalizable semantic segmentation of 3d scenes.
\newblock \emph{arXiv preprint arXiv:2111.13260}, 2021.

\bibitem[Vu et~al.(2022{\natexlab{a}})Vu, Kim, Luu, Nguyen, Kim, and Yoo]{vu2022softgroup++}
Thang Vu, Kookhoi Kim, Tung~M Luu, Thanh Nguyen, Junyeong Kim, and Chang~D Yoo.
\newblock Softgroup++: Scalable 3d instance segmentation with octree pyramid grouping.
\newblock \emph{arXiv preprint arXiv:2209.08263}, 2022{\natexlab{a}}.

\bibitem[Vu et~al.(2022{\natexlab{b}})Vu, Kim, Luu, Nguyen, and Yoo]{vu2022softgroup}
Thang Vu, Kookhoi Kim, Tung~M Luu, Thanh Nguyen, and Chang~D Yoo.
\newblock Softgroup for 3d instance segmentation on point clouds.
\newblock In \emph{Proceedings of the IEEE/CVF Conference on Computer Vision and Pattern Recognition}, pages 2708--2717, 2022{\natexlab{b}}.

\bibitem[Wang et~al.(2023)Wang, Vasu, Faghri, Vemulapalli, Farajtabar, Mehta, Rastegari, Tuzel, and Pouransari]{Wang2023SAMCLIPMV}
Haoxiang Wang, Pavan Kumar~Anasosalu Vasu, Fartash Faghri, Raviteja Vemulapalli, Mehrdad Farajtabar, Sachin Mehta, Mohammad Rastegari, Oncel Tuzel, and Hadi Pouransari.
\newblock Sam-clip: Merging vision foundation models towards semantic and spatial understanding.
\newblock \emph{ArXiv}, abs/2310.15308, 2023.

\bibitem[Wu et~al.(2023)Wu, Li, Ding, Li, Cheng, Tong, and Loy]{Wu2023BetrayedBC}
Jianzong Wu, Xiangtai Li, Henghui Ding, Xia Li, Guangliang Cheng, Yunhai Tong, and Chen~Change Loy.
\newblock Betrayed by captions: Joint caption grounding and generation for open vocabulary instance segmentation.
\newblock In \emph{Proceedings of the IEEE/CVF International Conference on Computer Vision}, pages 21938--21948, 2023.

\bibitem[Yamazaki et~al.(2023)Yamazaki, Hanyu, Vo, Pham, Tran, Doretto, Nguyen, and Le]{yamazaki2023open}
Kashu Yamazaki, Taisei Hanyu, Khoa Vo, Thang Pham, Minh Tran, Gianfranco Doretto, Anh Nguyen, and Ngan Le.
\newblock Open-fusion: Real-time open-vocabulary 3d mapping and queryable scene representation.
\newblock \emph{arXiv preprint arXiv:2310.03923}, 2023.

\bibitem[Yang et~al.(2023)Yang, Wu, He, Zhao, and Liu]{yang2023sam3d}
Yunhan Yang, Xiaoyang Wu, Tong He, Hengshuang Zhao, and Xihui Liu.
\newblock Sam3d: Segment anything in 3d scenes.
\newblock \emph{arXiv preprint arXiv:2306.03908}, 2023.

\bibitem[Yeshwanth et~al.(2023)Yeshwanth, Liu, Nie{\ss}ner, and Dai]{yeshwanth2023scannet++}
Chandan Yeshwanth, Yueh-Cheng Liu, Matthias Nie{\ss}ner, and Angela Dai.
\newblock Scannet++: A high-fidelity dataset of 3d indoor scenes.
\newblock In \emph{Proceedings of the IEEE/CVF International Conference on Computer Vision}, pages 12--22, 2023.

\bibitem[Yin et~al.(2023)Yin, Liu, Xiao, Cohen-Or, Huang, and Chen]{yin2023sai3d}
Yingda Yin, Yuzheng Liu, Yang Xiao, Daniel Cohen-Or, Jingwei Huang, and Baoquan Chen.
\newblock Sai3d: Segment any instance in 3d scenes.
\newblock \emph{arXiv preprint arXiv:2312.11557}, 2023.

\bibitem[Zhang et~al.(2020)Zhang, Zhu, Zheng, and Xu]{zhang2020fusion}
Jiazhao Zhang, Chenyang Zhu, Lintao Zheng, and Kai Xu.
\newblock Fusion-aware point convolution for online semantic 3d scene segmentation.
\newblock In \emph{Proceedings of the IEEE/CVF conference on computer vision and pattern recognition}, pages 4534--4543, 2020.

\bibitem[Zhang et~al.(2023)Zhang, Dai, Meng, Fan, Chen, Xu, and Wang]{zhang20233d}
Jiazhao Zhang, Liu Dai, Fanpeng Meng, Qingnan Fan, Xuelin Chen, Kai Xu, and He Wang.
\newblock 3d-aware object goal navigation via simultaneous exploration and identification.
\newblock In \emph{Proceedings of the IEEE/CVF Conference on Computer Vision and Pattern Recognition}, pages 6672--6682, 2023.

\bibitem[Zheng et~al.(2019)Zheng, Zhu, Zhang, Zhao, Huang, Niessner, and Xu]{zheng2019active}
Lintao Zheng, Chenyang Zhu, Jiazhao Zhang, Hang Zhao, Hui Huang, Matthias Niessner, and Kai Xu.
\newblock Active scene understanding via online semantic reconstruction.
\newblock In \emph{Computer Graphics Forum}, pages 103--114. Wiley Online Library, 2019.

\bibitem[Zhi et~al.(2021)Zhi, Laidlow, Leutenegger, and Davison]{zhi2021place}
Shuaifeng Zhi, Tristan Laidlow, Stefan Leutenegger, and Andrew~J Davison.
\newblock In-place scene labelling and understanding with implicit scene representation.
\newblock In \emph{Proceedings of the IEEE/CVF International Conference on Computer Vision}, pages 15838--15847, 2021.

\bibitem[Zhou et~al.(2022)Zhou, Girdhar, Joulin, Kr{\"a}henb{\"u}hl, and Misra]{zhou2022detecting}
Xingyi Zhou, Rohit Girdhar, Armand Joulin, Philipp Kr{\"a}henb{\"u}hl, and Ishan Misra.
\newblock Detecting twenty-thousand classes using image-level supervision.
\newblock In \emph{European Conference on Computer Vision}, pages 350--368. Springer, 2022.

\end{thebibliography}
}

\end{document}